\documentclass[a4paper,twoside]{article}

\usepackage{epsfig}
\usepackage{subcaption}
\usepackage{calc}
\usepackage{amssymb}
\usepackage{amstext}
\usepackage{amsmath}
\usepackage{amsthm}
\usepackage{multicol}
\usepackage{pslatex}
\usepackage{apalike}
\usepackage[bottom]{footmisc}
\usepackage{SCITEPRESS}     
\usepackage{xcolor}
\usepackage{hyperref}
\usepackage{url}
\usepackage{amsfonts}
\usepackage{algorithmic}
\usepackage{graphicx}
\usepackage{textcomp}
\usepackage{cite}
\def\BibTeX{{\rm B\kern-.05em{\sc i\kern-.025em b}\kern-.08em
    T\kern-.1667em\lower.7ex\hbox{E}\kern-.125emX}}

\begin{document}

\title{Machine Fault Classification using Hamiltonian Neural Networks}

\author{\authorname{Jeremy Shen\sup{1}, Jawad Chowdhury\sup{2}, Sourav Banerjee\sup{3} and Gabriel Terejanu\sup{2}}
\affiliation{\sup{1}Dept. of Electrical and Computer Engineering, University of Michigan, Ann Arbor, MI, USA}
\affiliation{\sup{2}Dept. of Computer Science, University of North Carolina at Charlotte, Charlotte, NC, USA}
\affiliation{\sup{3}Dept. of Mechanical Engineering, University of South Carolina, Columbia, SC, USA}
\email{jmshen@umich.edu, mchowdh5@uncc.edu, banerjes@cec.sc.edu, gabriel.terejanu@uncc.edu}
}


\abstract{A new approach is introduced to classify faults in rotating machinery based on the total energy signature estimated from sensor measurements. The overall goal is to go beyond using black-box models and incorporate additional physical constraints that govern the behavior of mechanical systems. Observational data is used to train Hamiltonian neural networks that describe the conserved energy of the system for normal and various abnormal regimes. The estimated total energy function, in the form of the weights of the Hamiltonian neural network, serves as the new feature vector to discriminate between the faults using off-the-shelf classification models. The experimental results are obtained using the MaFaulDa database, where the proposed model yields a promising area under the curve (AUC) of $0.78$ for the binary classification (normal vs abnormal) and $0.84$ for the multi-class problem (normal, and $5$ different abnormal regimes).}

\onecolumn \maketitle \normalsize \setcounter{footnote}{0} \vfill

\section{\uppercase{Introduction}}
\label{sec:introduction}
A cost-effective method for ensuring component reliability is to enhance the current schedule-based maintenance approach with deterministic component health and usage data to inform selective and targeted maintenance activities. Condition monitoring and fault diagnosis systems are required to guard against unexpected failures in safety-critical and production applications. Early fault detection can reduce unplanned failures, which will in turn reduce life cycle costs and increase readiness and mission assurances. Irrespective of different machinery, manufacturing tools like CNC machines, heavy equipment, aircraft, helicopters, space vehicles, car engines, and machines generate vibrations. The analysis of these vibration data is the key to detecting machinery degradation before the equipment or the structure fails. Machine faults usually leave key indications of its internal signature through the changes in modal parameters. For example, they may change the natural frequency of the system, generate unique damping characteristics, degradation in stiffness, generation of acoustic frequencies, etc. The defects and faults in the system may also generate a different form of energy transduction from mechanical to electrical or to electromagnetic energy, which leaves unique signatures. The statistical features of vibration signals in the time, frequency, and time-frequency domains each have different strengths for detecting fault patterns, which has been thoroughly studied~\cite{Nayana2017,Van2020,Li2016}. Various approaches have been proposed to extract features from these vibration signals using time-domain and frequency-domain analysis~\cite{lei2008new}, Fourier and wavelet transform~\cite{LI2013497}, and manifold learning~\cite{jiang2009machinery}. It is also shown that integration and hybridization of feature extraction algorithms can yield synergies that combine strengths and eliminate weaknesses~\cite{Usamentiaga2013,Rizzo2006}. Most of the work done in this area is based on data collected from vibration sensors~\cite{Yen2000,Seshadrinath2014,Bellini2008}, which are cheap and enable non-intrusive deployments. However, they generate huge datasets. As these data sets are very big in nature finding the above-mentioned unique features, and their respective paraxial contributions are extremely challenging. Hence, recently several feature extraction-driven machine learning algorithms are deployed to solve this challenge~\cite{Amin2015,Wei2019,Caggiano2019,Yin2019}.   

One of the challenges with building and deploying machine learning models to support decision-making is  achieving a level of generalization that allows us to learn on one part of the data distribution and predict on another. This challenge is amplified when learning using data from physical systems, as machine learning models such as neural networks (NN) capture an approximation of the underlying physical laws. Recently, new approaches have emerged under the umbrella of physics-informed neural networks (PINN)~\cite{karniadakis2021physics} to train NN that not only fit the observational data but also respect the underlying physics. This work leverages the Hamiltonian neural network (HNN)~\cite{Greydanus2019} to learn the Hamiltonian equations of energy-conserving dynamical systems from noisy data.

HNNs are used to characterize the total energy of rotating machinery, which is part of a wide range of applications such as power turbines, helicopters, and CNC machines just to name a few. In Ref.~\cite{Ribeiro2017}, the authors proposed a similarity-based model to calculate the similarity score of a signal with a set of prototype signals that characterize a target operating condition. These similarity score features are used in conjunction with time and spectral domain features to classify the behavior of the system using off-the-shelf classification models, such as random forests.

The main contribution of this work is to be intentional with respect to the underlying physics of the rotating machinery when generating discriminatory features. Namely, the conservation of energy is used as an inductive bias in the development and training of the HNN. While these mechanical systems are dissipative in nature, we assume that for short periods of time, the energy of the system is conserved due to the energy injected by the motor. The features derived by our approach are in the form of the weights of the HNN, which characterize the total energy of the system. In other words, we attempt to identify the operating regime based on the energy function. As with the previous approaches, these physics-informed features are then used to train off-the-shelf classifiers, such as logistic regressions and random forests to predict the condition of the mechanical system. The experimental results are performed on the Machinery Fault Database (MaFaulDa)\footnote{\scriptsize \url{http://www02.smt.ufrj.br/~offshore/mfs/page_01.html}} from the Federal University of Rio de Janeiro. The proposed system yields a promising area under the curve (AUC) of $0.78$ for both the binary classification (normal vs abnormal) and $0.84$ for the multi-class problem (normal, and $5$ different abnormal regimes).

This paper is structured as follows: Section~\ref{section:background} introduces the background on the HNN and MaFaulDa dataset. Section~\ref{section:methodology} presents our proposed approach to derive physics-informed features to classify operating conditions. Section~\ref{section:experiments} shows the empirical evaluations and Section~\ref{section:discussion} summarizes our findings.

\section{\uppercase{Background}}
\label{section:background}

\subsection{Hamiltonian Neural Networks}
The Hamiltonian equations of motion, Eq.~\ref{eq:Hamiltonian}, describe the mechanical system in terms of canonical coordinates, position $\mathbf{q}$ and momentum $\mathbf{p}$, and the Hamiltonian of the system $\mathcal{H}$.
\begin{equation}\label{eq:Hamiltonian}
    \frac{d\mathbf{q}}{dt} = \frac{\partial \mathcal{H}}{\partial \mathbf{p}}, \quad \frac{d\mathbf{p}}{dt} = - \frac{\partial \mathcal{H}}{\partial \mathbf{q}}
\end{equation}
Instead of using neural networks to directly learn the Hamiltonian vector field $\big( \frac{\partial \mathcal{H}}{\partial \mathbf{p}}, - \frac{\partial \mathcal{H}}{\partial \mathbf{q}} \big)$, the approach used by Hamiltonian neural networks is to learn a parametric function in the form of a neural network for the Hamiltonian itself~\cite{Greydanus2019}. This distinction accounts for learning the exact quantity of interest and it allows us to also easily obtain the vector field by taking the derivative with respect to the canonical coordinates via automatic differentiation. Given the training data, the parameters of the HNN are learned by minimizing the following loss function, Eq.~\ref{eq:loss}.
\begin{equation}\label{eq:loss}
    \mathcal{L} = \left\Vert \frac{\partial \mathcal{H}}{\partial \mathbf{p}} - \frac{d\mathbf{q}}{dt} \right\Vert_2 + \left\Vert \frac{\partial \mathcal{H}}{\partial \mathbf{q}} + \frac{d\mathbf{p}}{dt} \right\Vert_2
\end{equation}

\subsection{Machinery Fault Database (MaFaulDa)}

\begin{figure*}[htp]
   \centering
   \includegraphics[width=0.85\linewidth]{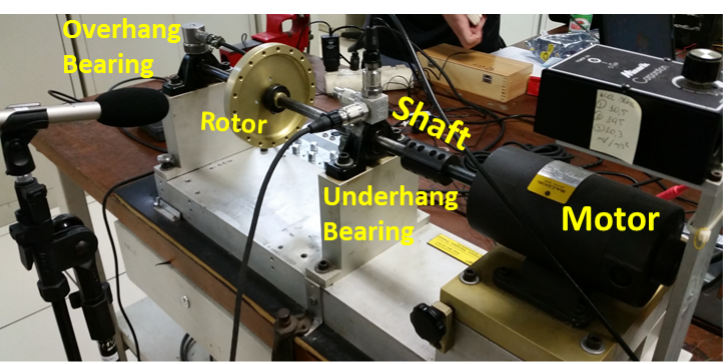}
   \caption{SpectraQuest System: Alignment-Balance-Vibration }
   \label{fig:spectraquest}
\end{figure*}

A comprehensive set of machine faults and vibration data was needed for the development and testing of the Hamiltonian-based feature extraction and classification of different operating states with damage/defects.  The Machinery Fault Database (MaFaulDa) consists of a comprehensive set of vibration data from a SpectraQuest Alignment-Balance-Vibration System, which includes multiple types of faults, see  Fig.~\ref{fig:spectraquest}. The equipment has two shaft-supporting bearings, a rotor, and a motor. Accelerometers are attached to the bearings to measure the vibration in the radial, axial, and tangential directions of each bearing. In addition, measurements from a tachometer (for measuring system rotation frequency) and a microphone (for capturing sound during system operation) are also included in the database. The database includes $10$ different operating states and a total of $1951$ sets of vibration data: (1) normal operation, (2) rotor imbalance, (3) underhang bearing fault: outer track, (4) underhang bearing fault: rolling elements, (5) underhang bearing fault: inner track, (6) overhang bearing fault: outer track, (7) overhang bearing fault: rolling elements, (8) overhang bearing fault: inner track, (9) horizontal shaft misalignment, (10) vertical shaft misalignment.
%

\textbf{Normal Operation}. There are $49$ sets of data from the system operating under normal conditions without any fault, each with a fixed rotating speed within the range from $737$ rpm to $3686$ rpm with steps of approximately $60$ rpm. 

\textbf{Rotor Imbalance}. To simulate different degrees of imbalanced operation, distinct loads of $(6$, $10$, $15$, $20$, $25$, $30$, $35)$ g were coupled to the rotor. The database includes a total of $333$ different imbalance-operation scenarios with combinations of loads and rotation frequencies.

\textbf{Bearing Faults}. As one of the most complex elements of the machine, the rolling bearings are the most susceptible elements to fault occurrence. Three defective bearings, each one with a distinct defective element (outer track, rolling elements, and inner track), were placed one at a time in each of the bearings. The three masses of $(6$, $10$, $20)$ g were also added to the rotor to induce a combination of rotor imbalance and bearing faults with various rotation frequencies. There is a total of $558$ underhang bearing fault scenarios and $513$ overhang bearing fault scenarios. 

\textbf{Horizontal Shaft Misalignment}. Horizontal shaft misalignment faults were induced by shifting the motor shaft horizontally of $(0.5$, $1.0$, $1.5$, $2.0)$ mm. The database includes a total of $197$ different scenarios with combinations of horizontal shaft misalignment and rotation frequencies. 

\textbf{Vertical Shaft Misalignment}. Vertical shaft misalignment faults were induced by shifting the motor shaft vertically of $(0.51$, $0.63$, $1.27$, $1.4$, $1.78$, $1.9)$ mm. The database includes a total of $301$ different scenarios with combinations of vertical shaft misalignment and rotation frequencies. 

\section{\uppercase{Methodology}}
\label{section:methodology}
\begin{figure*}[h]
   \centering
   \includegraphics[width=0.85\linewidth]{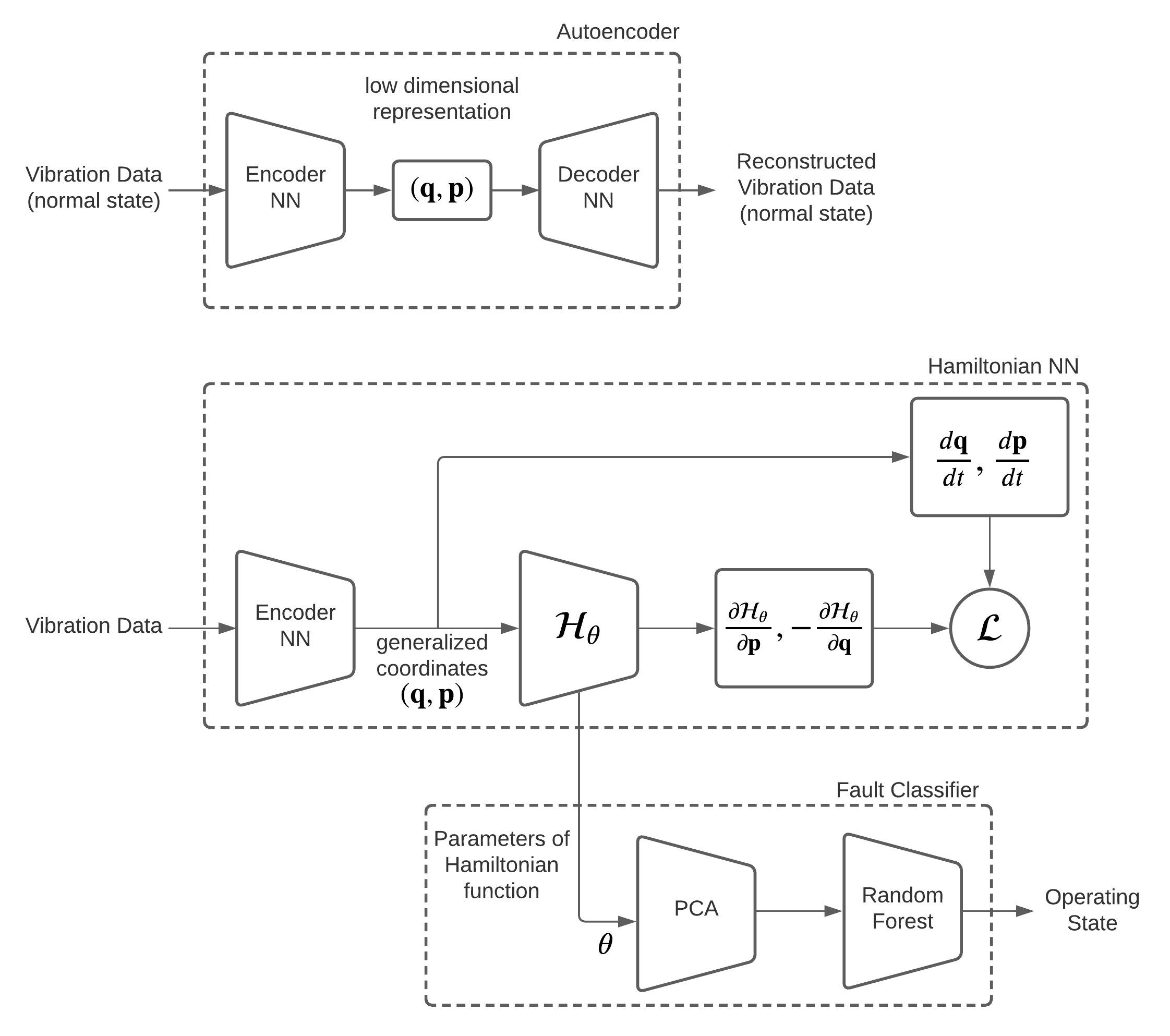}
   \caption{The proposed fault classification model}
   \label{fig:architecture}
\end{figure*}

The approach proposed to identify the operating state of the rotating machinery is to learn the total energy of the system from vibration data using HNN and use the parameters of the Hamiltonian as discriminating features. The intuition is that the total energy signature is different under various faults. The main assumption that we make is that the energy of the system is conserved for short periods of time thanks to the energy injected by the motor, which allows us to use the HNN~\cite{Greydanus2019}. The overall model architecture is shown in Fig.~\ref{fig:architecture}.

The first step is to develop a set of generalized coordinates from the raw vibration data using an autoencoder trained only on the data from normal conditions. The encoder NN is then used to generate a low dimensional representation ($2$D in our case) from the $8$ vibration measurements taken in any operating regime. This approach in developing arbitrary coordinates has been proposed in the original HNN paper~\cite{Greydanus2019}.

Using the newly developed coordinates, the second step is to train an HNN for each sequence of data generated at $50$ kHz sampling rate during $5$ s. The parameters $\theta$ of the Hamiltonian $\mathcal{H}_\theta$ fully characterize the energy function of the operating state and they can be used to train a classifier.

The parameterization of the Hamiltonian is high-dimensional ($41,200$ weights in our case) as it depends on the number of layers and hidden neurons per layer chosen in HNN. As a result, we have chosen to reduce its dimension using principal component analysis (PCA) before training a classifier as a random forest in the last modeling step.

\section{\uppercase{Numerical Results}}
\label{section:experiments}

The proposed fault classification system has been used with the MaFaulDa dataset and a $70:30$ split into train and test data. Given the imbalance of collected data, namely $49$ datasets recorded for normally operating motors vs. 1800+ datasets recorded for all faulty operating motors, we have used the synthetic minority over-sampling technique (SMOTE)~\cite{chawla2002smote} to create synthetic data points for the minority class. The PyCaret\footnote{\url{https://pycaret.org}} framework was used to develop the classifiers and preprocess the HNN features.

Two different tasks are considered. The first is the binary classification where we are discriminating between normal and abnormal conditions using a random forest, and the second is the multi-class problem where we are discriminating using a logistic regression between the classes listed in Table~\ref{tbl:one_vs_rest}, where class $0$ is the normal regime.

\begin{table}[htp]
\caption{Pairwise classification - results on testing set}
\label{tbl:one_vs_rest}
\centering
\begin{tabular}{|l|l|l|l|}
\hline
Class & \textbf{normal vs X} & \textbf{AUC} & \textbf{F1-score} \\ \hline
1 & horizontal-misalign. & 0.59 & 0.80 \\ \hline
2 & imbalance & 0.92 & 0.95 \\ \hline
3 & overhang & 0.85 & 0.85 \\ \hline
4 & underhang & 0.80 & 0.88 \\ \hline
5 & vertical-misalign. & 0.91 & 0.92 \\ \hline
\end{tabular}
\end{table}

\begin{figure}[htp]
   \centering
   \includegraphics[width=0.98\linewidth]{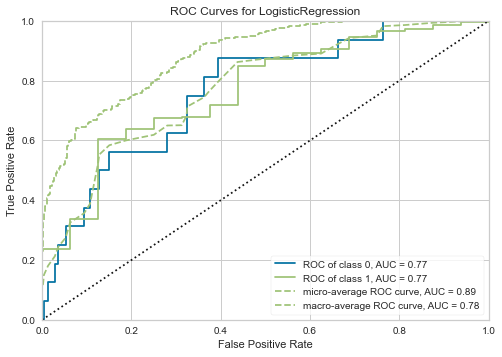}
   \caption{Results on test set - binary classification}
   \label{fig:ROC_binary}
   \vspace{0.3in}
\end{figure}

\begin{figure}[htp]
   \centering
   \includegraphics[width=0.98\linewidth]{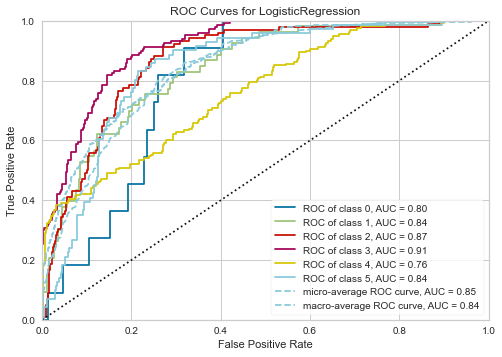}
   \caption{Results on test set - multi-class problem}
   \label{fig:ROC_multiclass}
   \vspace{0.3in}
\end{figure}

The receiver operating characteristic (ROC) curves are provided for both tasks in Figs.~\ref{fig:ROC_binary} and \ref{fig:ROC_multiclass} respectively. The macro-averaged AUC calculated on the test data is $0.78$ for the binary classification and $0.84$ for the multi-class problem and the F1 score is $0.96$ and $0.51$ respectively, which demonstrates the viability of physics-informed features from HNN to capture the state of the system. These classification problems are imbalanced due to the skewed distribution of examples across the classes and as a result, we have chosen not to report the accuracy as it was reported in prior work~\cite{Ribeiro2017,MARINS20181913}. We note however that Ref.\cite{Ribeiro2017} reports an F1-score of $0.99$ on a $10$-fold cross-validation exercise, which is higher that the $0.96$ on our binary classification, and that our multiclass F1-score is lower due to the aggregation of bearing fault classes.

\begin{figure}[htp]
   \centering
   \vspace{-0.25in}
   \includegraphics[width=1.15\linewidth]{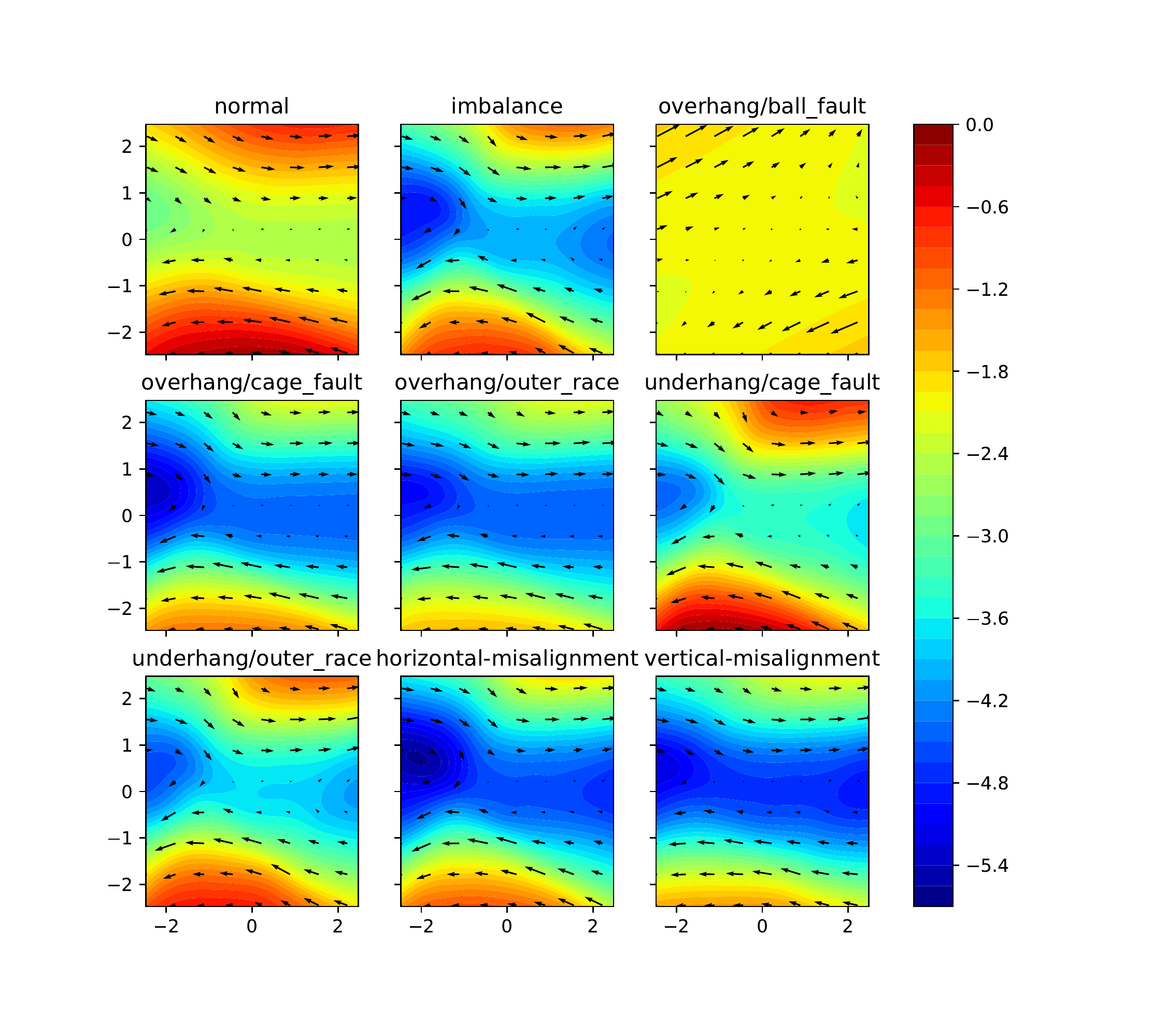}
  \vspace{-0.2in}
   \caption{Phase portraits of various operating conditions}
   \label{fig:diff_regimes}
   \vspace{0.2in}
\end{figure}

Table~\ref{tbl:one_vs_rest} shows the AUC for pairwise classification between each unique defective operating condition and the normal condition. Fig.~\ref{fig:diff_regimes} shows the phase spaces of $10$ different operating conditions ($1$ normal and $9$ faulty). Interestingly, among all the pairwise comparisons, the model finds the discrimination between normal and horizontal-misalignment regimes rather challenging, which we plan to further explore in future studies. We do expect that the faults introduced generate slight changes in the phase portraits of various regimes, see Fig.~\ref{fig:diff_regimes}. However, we find qualitatively that the phase portrait of overhang/ball-fault is significantly different than the rest, which suggests that the sub-classes of overhang, namely ball-fault, cage-fault, and outer-race should be treated as classes on their own.

\begin{figure*}[htp]
   \centering
   \vspace{-0.25in}
   \includegraphics[width=0.80\linewidth]{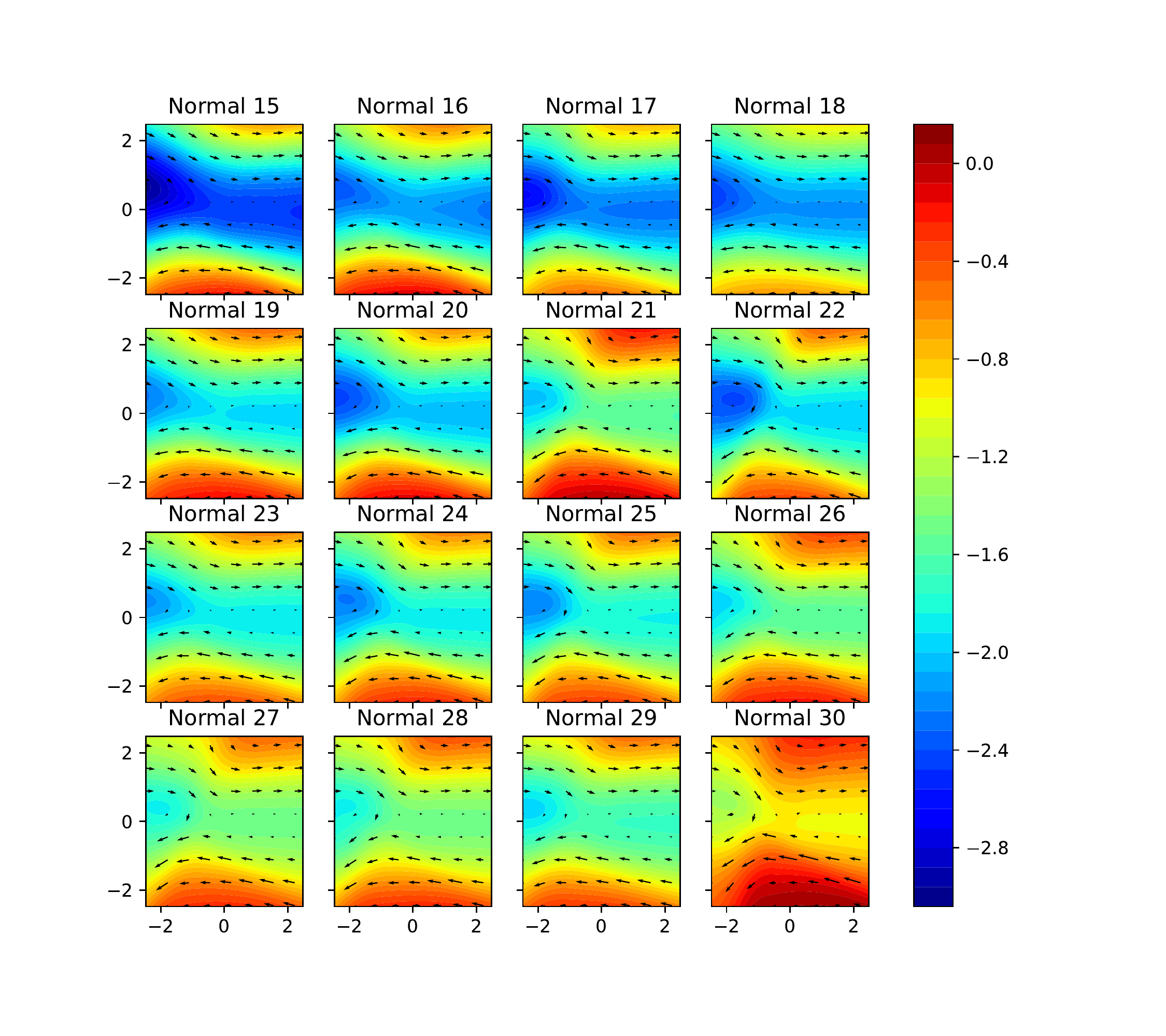}
  \vspace{-0.2in}
   \caption{The effect of rotation frequency on the Hamiltonian}
   \label{fig:normal_behaviour}
\end{figure*} 

\textbf{Discussion on the effect of rotation frequency on the Hamiltonian}. Fig.~\ref{fig:normal_behaviour} shows the Hamiltonian of normally operating motors operating at different speeds. Interestingly, even though this has not been enforced, the general structure of the Hamiltonian vector field remains largely the same across various speeds, while the magnitude of the Hamiltonian increases at higher speeds as expected. It can be concluded in this case that the vector field is dependent on the operating condition, and the magnitude is dependent on the operating speed.


\begin{figure}[htp]
   \centering
   \includegraphics[width=1.05\linewidth]{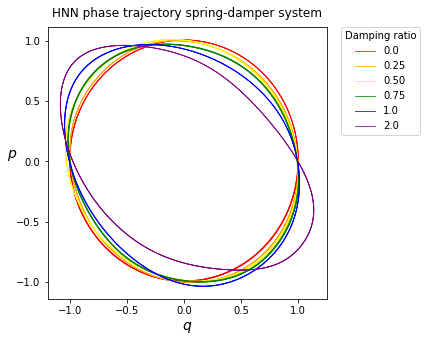}
   \caption{HNN results on variable damping ratios}
   \label{fig:damping_variations}
   \vspace{0.43in}
\end{figure}

\textbf{Discussion on HNN on Dissipative Systems}. The HNN has the ability to learn the total energy of a number of systems~\cite{Greydanus2019}, including an ideal mass-spring system. Although the HNN is designed to conserve energy, it is interesting to consider what the HNN learns from dissipative systems. We believe that the methodology is more broadly applicable and it applies also when this assumption does not hold.

We have used a mass-spring-damper to experiment with the behavior of HNN for dissipative systems. This is a non-conservative system and the Hamiltonian formulated by the HNN is not a conventional solution to the mass-spring-damper system, the conserved quantity is not the total energy, and the generalized coordinates are not position and momentum defined by classical mechanics. Nevertheless, a qualitative analysis of the trajectories shows that the HNN creates unique solutions for each value of the damping ratio, see Fig.~\ref{fig:damping_variations}. While we are unable to use conventional physics to understand the results of the HNN on the mass-spring-damper system, it is evident that the results can be used to discriminate between the different systems. 

\vspace{-0.2in}
\section{\uppercase{Conclusions}}
\label{section:discussion}
A novel predictive model is introduced to discriminate between normal and abnormal operating regimes of rotating machinery. The model is based on the total energy signature of the system learned using a Hamiltonian Neural Network. The performance measures obtained from the experimental data suggest that the proposed physics-informed features are an excellent candidate for machine fault classification.


\section*{Acknowledgement}
Research was sponsored by the National Institute of Food and Agriculture under Grant Number 2017-67017-26167 and by
the Army Research Office under Grant Number W911NF-22-1-0035. The views and conclusions contained in this document are those of the authors and should not be interpreted as representing the official policies, either
expressed or implied, of the Army Research Office, the National Institute of Food and Agriculture, or the U.S. Government. The U.S. Government is
authorized to reproduce and distribute reprints for Government purposes notwithstanding any copyright
notation herein.
\newpage
\bibliographystyle{apalike}
{\small
\bibliography{main}}

\begin{thebibliography}{}

\bibitem[Amin et~al., 2015]{Amin2015}
Amin, H.~U., Malik, A.~S., Ahmad, R.~F., Badruddin, N., Kamel, N., Hussain, M.,
  and Chooi, W.-T. (2015).
\newblock Feature extraction and classification for eeg signals using wavelet
  transform and machine learning techniques.
\newblock {\em Australasian Physical \& Engineering Sciences in Medicine},
  38(1):139--149.

\bibitem[Bellini et~al., 2008]{Bellini2008}
Bellini, A., Immovilli, F., Rubini, R., and Tassoni, C. (2008).
\newblock Diagnosis of bearing faults of induction machines by vibration or
  current signals: A critical comparison.
\newblock In {\em 2008 IEEE Industry Applications Society Annual Meeting},
  pages 1--8.

\bibitem[Caggiano et~al., 2019]{Caggiano2019}
Caggiano, A., Zhang, J., Alfieri, V., Caiazzo, F., Gao, R., and Teti, R.
  (2019).
\newblock Machine learning-based image processing for on-line defect
  recognition in additive manufacturing.
\newblock {\em CIRP Annals}, 68(1):451--454.

\bibitem[Chawla et~al., 2002]{chawla2002smote}
Chawla, N.~V., Bowyer, K.~W., Hall, L.~O., and Kegelmeyer, W.~P. (2002).
\newblock Smote: synthetic minority over-sampling technique.
\newblock {\em Journal of artificial intelligence research}, 16:321--357.

\bibitem[Greydanus et~al., 2019]{Greydanus2019}
Greydanus, S.~J., Dzumba, M., and Yosinski, J. (2019).
\newblock Hamiltonian neural networks.
\newblock In {\em 33rd Conference on Neural Information Processing Systems
  (NeurIPS), Vancouver, Canada}.

\bibitem[Jiang et~al., 2009]{jiang2009machinery}
Jiang, Q., Jia, M., Hu, J., and Xu, F. (2009).
\newblock Machinery fault diagnosis using supervised manifold learning.
\newblock {\em Mechanical systems and signal processing}, 23(7):2301--2311.

\bibitem[Karniadakis et~al., 2021]{karniadakis2021physics}
Karniadakis, G.~E., Kevrekidis, I.~G., Lu, L., Perdikaris, P., Wang, S., and
  Yang, L. (2021).
\newblock Physics-informed machine learning.
\newblock {\em Nature Reviews Physics}, 3(6):422--440.

\bibitem[Lei et~al., 2008]{lei2008new}
Lei, Y., He, Z., and Zi, Y. (2008).
\newblock A new approach to intelligent fault diagnosis of rotating machinery.
\newblock {\em Expert Systems with applications}, 35(4):1593--1600.

\bibitem[Li et~al., 2016]{Li2016}
Li, C., SÃ¡nchez, R.-V., Zurita, G., Cerrada, M., and Cabrera, D. (2016).
\newblock Fault diagnosis for rotating machinery using vibration measurement
  deep statistical feature learning.
\newblock {\em Sensors}, 16(6).

\bibitem[Li et~al., 2013]{LI2013497}
Li, P., Kong, F., He, Q., and Liu, Y. (2013).
\newblock Multiscale slope feature extraction for rotating machinery fault
  diagnosis using wavelet analysis.
\newblock {\em Measurement}, 46(1):497--505.

\bibitem[Marins et~al., 2018]{MARINS20181913}
Marins, M.~A., Ribeiro, F.~M., Netto, S.~L., and {da Silva}, E.~A. (2018).
\newblock Improved similarity-based modeling for the classification of
  rotating-machine failures.
\newblock {\em Journal of the Franklin Institute}, 355(4):1913--1930.
\newblock Special Issue on Recent advances in machine learning for signal
  analysis and processing.

\bibitem[Nayana and Geethanjali, 2017]{Nayana2017}
Nayana, B.~R. and Geethanjali, P. (2017).
\newblock Analysis of statistical time-domain features effectiveness in
  identification of bearing faults from vibration signal.
\newblock {\em IEEE Sensors Journal}, 17(17):5618--5625.

\bibitem[Ribeiro et~al., 2017]{Ribeiro2017}
Ribeiro, F., Marins, M., Netto, S., and da~Silva, E. (2017).
\newblock Rotating machinery fault diagnosis using similarity-based models.
\newblock In {\em XXXV Simpósio Brasileiro de Telecomunicações e
  Processamento de Sinais, São Pedro, Brasil}.

\bibitem[Rizzo and di~Scalea, 2006]{Rizzo2006}
Rizzo, P. and di~Scalea, F.~L. (2006).
\newblock Feature extraction for defect detection in strands by guided
  ultrasonic waves.
\newblock {\em Structural Health Monitoring}, 5(3):297--308.

\bibitem[Seshadrinath et~al., 2014]{Seshadrinath2014}
Seshadrinath, J., Singh, B., and Panigrahi, B.~K. (2014).
\newblock Investigation of vibration signatures for multiple fault diagnosis in
  variable frequency drives using complex wavelets.
\newblock {\em IEEE Transactions on Power Electronics}, 29(2):936--945.

\bibitem[Usamentiaga et~al., 2013]{Usamentiaga2013}
Usamentiaga, R., Venegas, P., Guerediaga, J., Vega, L., and L\'{o}pez, I.
  (2013).
\newblock Feature extraction and analysis for automatic characterization of
  impact damage in carbon fiber composites using active thermography.
\newblock {\em NDT \& E International}, 54:123--132.

\bibitem[Van et~al., 2020]{Van2020}
Van, B., Van~Hoa, N., Nguyen, H., and Jang, Y.~M. (2020).
\newblock Statistical feature extraction in machine fault detection using
  vibration signal.
\newblock In {\em International Conference on Information and Communication
  Technology Convergence (ICTC)}, pages 666--669.

\bibitem[Wei et~al., 2019]{Wei2019}
Wei, J., Chu, X., Sun, X.-Y., Xu, K., Deng, H.-X., Chen, J., Wei, Z., and Lei,
  M. (2019).
\newblock Machine learning in materials science.
\newblock {\em InfoMat}, 1(3):338--358.

\bibitem[Yen and Lin, 2000]{Yen2000}
Yen, G. and Lin, K.-C. (2000).
\newblock Wavelet packet feature extraction for vibration monitoring.
\newblock {\em IEEE Transactions on Industrial Electronics}, 47(3):650--667.

\bibitem[Yin et~al., 2019]{Yin2019}
Yin, L., Ye, B., Zhang, Z., Tao, Y., Xu, H., {Salas Avila}, J.~R., and Yin, W.
  (2019).
\newblock A novel feature extraction method of eddy current testing for defect
  detection based on machine learning.
\newblock {\em NDT \& E International}, 107:102108.

\end{thebibliography}

\end{document}